\documentclass[sigconf]{acmart}
\AtBeginDocument{%
  }

\setcopyright{acmcopyright}
\acmDOI{10.1145/nnnnnnn.nnnnnnn} 
\acmISBN{978-x-xxxx-xxxx-x/YY/MM} 
\acmConference[GECCO '26]{The Genetic and Evolutionary Computation Conference 2026}{July 13--17, 2026}{San José, Costa Rica}
\acmYear{2026}
\copyrightyear{2026}

\usepackage{graphicx}
\usepackage{amsmath}
\usepackage{booktabs}
\usepackage[table]{xcolor} 
\usepackage{float}
\usepackage[ruled,linesnumbered]{algorithm2e}

\hypersetup{
    colorlinks=true,
    linkcolor=blue,
    urlcolor=blue,
    citecolor=blue
}

\title{Breaking Validity-Induced Boundaries to Expand Algorithm Search Space: A Two-Stage AST-Based Operator for LLM-Driven Automated Heuristic Evolution}

\author{Shengming Sun}
\affiliation{%
  \institution{School of Mathematics and Statistics, Xi'an Jiaotong University}
  \city{Xi'an}
  \country{China}
}
\email{3245814902@stu.xjtu.edu.cn}

\author{Jialong Shi}
\authornote{$\dagger$ Corresponding author.}
\affiliation{%
  \institution{School of Mathematics and Statistics, Xi'an Jiaotong University}
  \city{Xi'an}
  \country{China}
}
\email{jialong.shi@xjtu.edu.cn}

\begin{document}

\begin{abstract}
Large Language Model (LLM) based automated heuristic design (AHD) has shown great potential in discovering efficient heuristics. Most existing LLM-AHD frameworks use semantic evolutionary operators that rely entirely on the LLM's pre-trained knowledge. These one-stage methods strictly require the generated code to be valid during the operation and often rely on a ``thought-code'' representation. We argue that this end-to-end generation fundamentally limits the exploration ability within the algorithm search space.

In this paper, we propose a two-stage, structure-based evolutionary operator for LLM-AHD. In the first stage, our approach directly performs crossover and mutation on the Abstract Syntax Trees (ASTs) of the heuristic code, intentionally generating diverse but often invalid structural variants. In the second stage, the LLM is employed to repair these invalid heuristics into executable, high-quality code. Depending on the underlying framework, either the raw invalid variants or the repaired heuristics are integrated into the population to preserve potential structural patterns. We demonstrate that the proposed operator can significantly enhance the search ability of state-of-the-art LLM-AHD algorithms, such as EoH-S. Experimental results on the Traveling Salesman Problem (TSP) and the Online Bin Packing Problem (OBP) show that our method effectively improves both optimization performance and convergence speed.
\end{abstract}



\maketitle

\section{INTRODUCTION}

Combinatorial optimization problems aim to find optimal solutions in discrete search spaces, and most are NP-hard \cite{Garey1979Computers}. Heuristic algorithms are widely used to solve such problems, but traditional design relies heavily on expert experience and suffers from low efficiency. In recent years, with the rapid development of large language models (LLMs), automated heuristic design (AHD) based on LLMs has made remarkable progress. Several evolutionary methods, such as EoH \cite{Liu2024Evolution}, EoH-S \cite{Liu2026Eoh-s}, MEoH \cite{Yao2025Multi-objective}, and ReEvo \cite{Ye2024Reevo}, have achieved strong performance on various optimization tasks. Meanwhile, some studies have explored how prompt engineering affects the effectiveness of LLM-based automatic algorithm design \cite{Loss2025From}.

However, existing LLM-driven AHD methods face critical limitations due to their evolutionary mechanisms. First, most current evolutionary operators are semantic-based and rely heavily on the LLMs' pre-trained knowledge \cite{Gurkan2025Lear,Bhaskar2024Theheuristic}. This reliance makes it difficult to discover novel heuristic strategies that go beyond the models' learned patterns. Second, these one-stage methods strictly require the heuristic code to remain valid and executable throughout the entire operation process. By forcing every intermediate generation to be a valid program --- often through a rigid ``thought-code'' representation --- these frameworks artificially restrict the available search space. Together, the over-reliance on pre-trained semantics and the strict requirement for continuous code validity severely limit the exploration ability within the broader algorithm space, often leading to limited diversity and premature convergence.

To overcome these limitations, this paper proposes an evolutionary operator framework based on Abstract Syntax Trees (ASTs) \cite{Sun2023Abstract}. In the first stage, we convert heuristic code into ASTs and directly perform structural crossover and mutation. This step intentionally generates invalid heuristic code variants (I-Codes) to break the LLM's pre-trained knowledge boundaries. In the second stage, the LLM is employed purely to repair these invalid heuristics back into executable code. By explicitly separating structural perturbation from semantic repair, our method effectively expands the search space and improves the overall exploration ability.

Our main contributions are as follows:
\begin{itemize}
    \item As an exploratory study on breaking the knowledge boundaries of LLMs, we propose a two-stage, AST-based evolutionary operator. This operator is designed not to replace existing semantic operators, but to complement them by enabling structural modifications that effectively expand the algorithm search space.
    \item We demonstrate the flexibility of the proposed operator by integrating it as a complementary enhancement into three representative LLM-AHD frameworks: EoH, ReEvo, and EoH-S.
    \item We conduct extensive experiments on typical combinatorial optimization problems (TSP and OBP). The results show that the enhanced methods outperform the original versions in optimization performance and convergence speed.
\end{itemize}

\section{RELATED WORK}

\textbf{LLM-driven Automatic Heuristic Design:}\quad The integration of large language models and evolutionary computation has become a key paradigm in automated heuristic design. Early work such as FunSearch revealed the capability of LLMs in program search and mathematical discovery \cite{Romera2024Mathematical}. Subsequently, Evolution of Heuristics (EoH) combined LLMs with evolutionary algorithms using text-level evolutionary operators for iterative heuristic generation and refinement \cite{Liu2024Evolution}, while ReEvo enhanced this process with short-term and long-term reflection mechanisms \cite{Ye2024Reevo}, and EoH-S further extended the framework to diverse and complementary heuristic set design \cite{Liu2026Eoh-s}. More recently, Chen et al. proposed a Tuning-Integrated Dynamic Evolution framework that decouples structural reasoning from parameter optimization to mitigate premature convergence and improve generated heuristics' performance \cite{Chen2026TIDE}. Collectively, these methods establish LLM-driven evolutionary computation as a highly effective approach for automated algorithm design.

\noindent\textbf{Abstract Syntax Trees (ASTs):} \quad Abstract Syntax Trees (ASTs) have been widely adopted in program analysis, code generation, and automated program repair \cite{Sun2023Abstract}. By representing code in a tree structure, ASTs enable robust and syntactically consistent structural modifications. Compared with naive text editing, AST-based operations preserve structural logic while allowing meaningful perturbations, making them ideal for generating diverse and structurally valid variants \cite{Song2024Revisiting}. Such properties have motivated their use in code evolution and program synthesis.

\noindent\textbf{Prompt Engineering Enhancing LLMs:}\quad Prompt engineering is the process of structurally processing inputs, which plays a crucial role in automated heuristic design based on large language models \cite{Chen2025Unleashing}. Studies have shown that many large language models are highly sensitive to prompts, with their performance varying significantly with different prompt templates \cite{He2024Doesprompt}. As a successful application example, the HiFo prompt employs two collaborative prompt strategies---prospection and hindsight---to guide large language models, enabling the generation of high-quality heuristics and significantly accelerating the convergence speed \cite{Chen2026Hifo-prompt}. Recently, several studies have also indicated that the code description component in prompts plays the most crucial role in guiding large language models to generate novel heuristics \cite{Huang2026From}.

\begin{figure*}[htbp]
    \centering
    \includegraphics[width=1\linewidth, alt={Three-stage AST-based operator framework: converting heuristics to ASTs, performing structural crossover/mutation, and repairing via LLM interaction}]{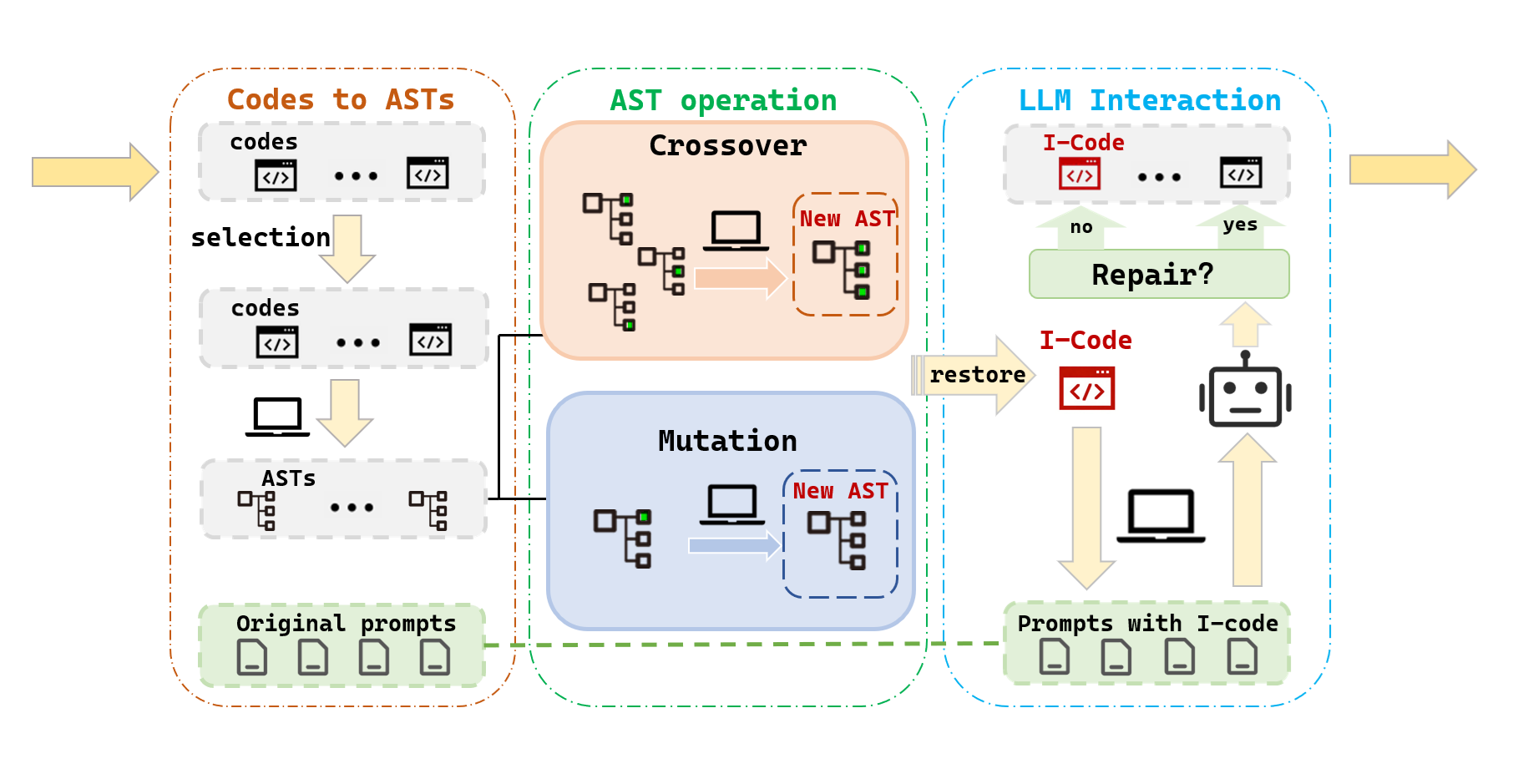}
    \caption{Framework of the AST-based operator. It comprises three core processes: (1) \textbf{Heuristics to ASTs}: Selected heuristics are converted into semantic trees via syntax tree manipulation. (2) \textbf{AST Operation}: Tree crossover and mutation are executed on the AST structure. (3) \textbf{LLM Interaction}: The new AST is converted into source code, integrated into the prompt, and supplied to the LLM to generate novel heuristics.}
    \label{fig:AST-P_framework}
\end{figure*}

\section{METHODOLOGY}
In this section, we present our AST-based operator, a novel evolutionary operator based on Abstract Syntax Trees that enables large language models (LLMs) to transcend the limitations of their inherent knowledge. The AST-based operator primarily consists of three components: converting codes into ASTs, performing operations on the trees, and interacting with LLMs to generate new codes. Additionally, this section demonstrates the application of the proposed operator in specific evolutionary frameworks (EoH, EoH-S, ReEvo). The framework of the proposed AST-based operator is shown in Figure~\ref{fig:AST-P_framework}.

\subsection{Design Rationale of the AST-based Operator}
\textbf{Invalid Code:} Research has shown that code description plays the most critical role in prompts. In LLM-driven Automatic Heuristic Design (AHD), the LLM only generates executable code, meaning that the search space of the LLM is limited to executable code alone. However, some invalid codes may provide new insights for the LLM, which we refer to as I-Codes (invalid codes).

Although invalid code cannot be executed, it may still be valuable for algorithm design. If we define the algorithm space $H$ as the set of strings capable of representing algorithmic ideas, and denote the set of executable code as $R$ and the set of invalid code as $I$, then $I$ is much more densely distributed than $R$ in this discrete space, since randomly altering a single character in any executable code will almost certainly result in invalid code. Under ideal conditions, all code generated by large language models is executable, which restricts the search space of heuristic automatic algorithm design to a subset of $R$. Therefore, we aim to expand the search space by generating invalid code via AST and then repairing such invalid code into executable code using large language models.

\begin{figure*}[htbp]
    \centering
    \includegraphics[width=0.8\linewidth, alt={Algorithm space partitioned into valid and invalid regions; AST operator bridges disjoint valid regions by traversing invalid space}]{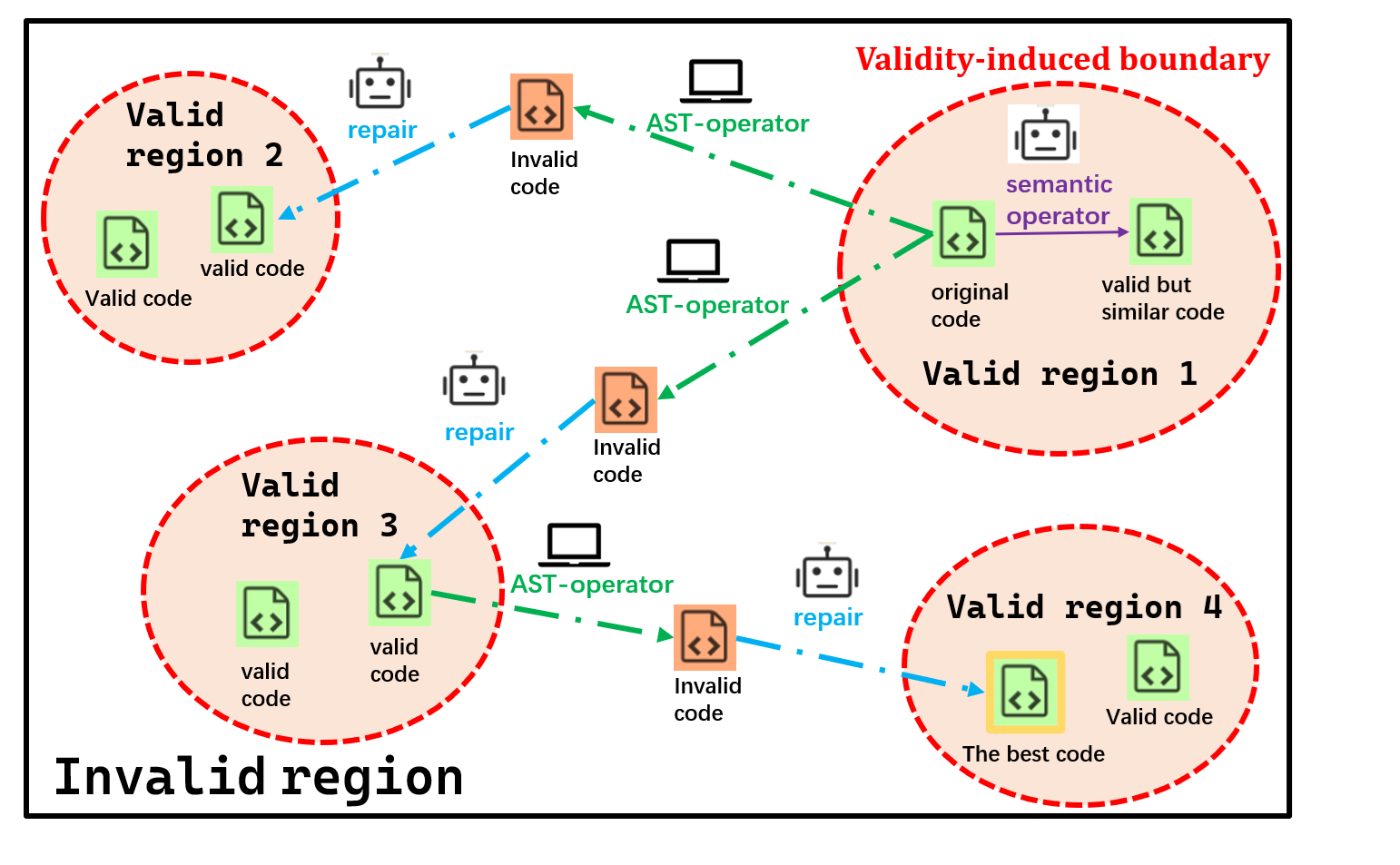}
    \caption{Illustration of search space expansion via I-Codes. Black rectangles represent the entire algorithm space; light orange ellipses denote valid regions; blank areas represent invalid regions. The AST operator bridges disjoint valid regions by traversing the invalid space, which semantic operators cannot cross.}
    \label{fig:placeholder}
\end{figure*}

In Figure~\ref{fig:placeholder}, we illustrate the ideal scenario with an example. There exist a valid region (containing valid code) and an invalid region (containing invalid code) in the algorithm space, and their relationship is roughly represented by light orange ellipses inside black rectangles. A valid code can be transformed into an invalid code (I-Code) through an AST operator, and the I-code can be converted into a valid code via repair. It is assumed that we only have the original code at the beginning. Starting from the original code, the algorithm applies an AST operator and a repair process to reach valid region 2 and find a valid code. Then, starting again from the original code, it applies a different AST operator and repair process to reach valid region 3 and find a new valid code. Furthermore, starting from the new valid code, it applies an AST operator and repair process to find the best code. However, if only semantic operators are used and I-code is not accepted, the algorithm cannot break through the validity-induced boundary and can only find valid but similar code.

\noindent\textbf{AST (Abstract Syntax Tree):} An Abstract Syntax Tree (AST) is a hierarchical tree-structured representation generated from source code through lexical analysis and syntactic parsing. Each node in the AST corresponds to a syntactic element such as a statement, expression, variable, operator, or control structure, while the edges represent grammatical dependencies between components \cite{Cui2010Code}. Compared with plain text, ASTs explicitly preserve the logical structure and grammatical rules of code, enabling systematic and structural modifications rather than arbitrary string edits. Based on this structural representation, we perform mechanical and controllable evolutionary operations on heuristic code:
\begin{itemize}
    \item  \textbf{Crossover}: We randomly select one subtree from each AST and exchange them to produce new hybrid structures. This operation combines logical components from different heuristics, creating novel code patterns that text-level methods cannot generate \cite{Koza1994Genetic}.
    \item \textbf{Mutation}: We randomly select a non-root node from the AST and delete it directly, which breaks the original logic and generates structurally perturbed variants \cite{Koza1994Genetic}.
\end{itemize}
Since these operations modify code at the structural level, the generated ASTs are not guaranteed to be syntactically correct. After performing crossover and mutation, we traverse the modified ASTs and convert them back into string-form code. The resulting code is allowed to be invalid (I-Code), which helps break the knowledge constraints of LLMs and expand the search space of automated heuristic design. Specifically, within the framework of automated heuristic design, the operator is completed in two steps: AST-based Destruction and LLM-based repair.

Algorithm~\ref{alg:AST_Destruct} presents the AST-based destruction: converting heuristics into Abstract Syntax Trees (ASTs), then disrupting them through crossover and mutation operations, followed by transforming them into invalid codes (I-Codes).

\begin{algorithm}[t]
\SetAlgoLined
\KwData{Valid codes $C_1$, $C_2$}
\KwResult{Invalid codes $C_{\text{del}}$, $C_{\text{cross}}$}
\tcp{Parse code into Abstract Syntax Trees}
$T_1 \leftarrow \text{CodeToAST}(C_1)$\;
$T_2 \leftarrow \text{CodeToAST}(C_2)$\;
\tcp{Random node deletion (destruction)}
$T_{\text{del}} \leftarrow \text{RandomDeleteNode}(T_1)$\;
\tcp{Random subtree crossover (destruction)}
$T_{\text{cross}} \leftarrow \text{RandomCrossover}(T_1, T_2)$\;
\tcp{Convert AST back to code (may produce invalid code)}
$C_{\text{del}} \leftarrow \text{ASTToCode}(T_{\text{del}})$\;
$C_{\text{cross}} \leftarrow \text{ASTToCode}(T_{\text{cross}})$\;
\Return{$C_{\text{del}}$, $C_{\text{cross}}$}
\caption{AST-Based Destruction Operators}
\label{alg:AST_Destruct}
\end{algorithm}

Algorithm~\ref{alg:LLM_Repair} illustrates the second step of the operator: first, determining whether the I-Code needs to be repaired; if not, directly adding the I-Code to the population; if repair is required, generating a repair prompt by incorporating the I-Code as needed to guide large language models (LLMs) in generating new heuristics, which are then added to the population.

\begin{algorithm}[t]
\SetAlgoLined
\KwData{Invalid Code (I-Code), Population, Large Language Models (LLMs)}
\KwResult{Updated Population}
\tcp{Judge whether the I-Code needs to be repaired}
$needRepair \leftarrow \text{RepairNeed}()$\;
\If{$\text{needRepair} = \text{false}$}
{
    \tcp{Add I-Code directly to the population}
    $\text{AddtoPopulation}(Population, I\text{-}Code)$\;
}
\Else
{
    \tcp{Generate repair prompt incorporating I-Code}
    $repairPrompt \leftarrow \text{GenerateRepairPrompt}(I\text{-}Code)$\;
    \tcp{Guide LLMs to generate new heuristics}
    $newCodes \leftarrow \text{LLMGenerate}(repairPrompt, LLMs)$\;
    \tcp{Add new codes to the population}
    $\text{AddtoPopulation}(Population, newCodes)$\;
}
\Return{$Population$}
\caption{LLM-Based Repair Operators}
\label{alg:LLM_Repair}
\end{algorithm}

\subsection{Application in EoH}
\textbf{Evolution of Heuristics (EoH)} is an automated heuristic design algorithm that integrates large language models and evolutionary algorithms with five evolutionary operators \cite{Liu2024Evolution}. To strengthen its evolutionary capacity, we redesign these operators using an AST-based approach. We introduce invalid heuristics into the evolutionary population and define a suitable fitness function to evaluate individuals during iteration.

\subsubsection{Redesigned Operators}
EoH utilizes five \textbf{evolutionary operators (E1, E2, M1, M2, M3)}; in this work, we redesign these operators on the basis of Abstract Syntax Trees (ASTs). The redesigned operators are classified into three categories according to whether the operation objects and generated heuristics are invalid heuristics.

\textbf{Valid Codes to Invalid Codes (VI):} Heuristic code is first converted into an abstract syntax tree, on which structural crossover and mutation are performed. The modified AST is then converted back to code, which may be invalid, and is directly inserted into the population as an \textbf{invalid code}.

\textbf{Invalid Codes to Valid Codes (IV):} Invalid codes are incorporated into the prompt, requiring the LLM to either repair them or generate novel heuristics distinct from the original ones, which are then added to the population as new \textbf{codes}.

\textbf{Invalid Codes to Invalid Codes (II):} Two strategies are adopted: (1) transforming invalid codes into ASTs for structural operations, then restoring them to code and directly adding them to the population as \textbf{invalid codes}; (2) invoking the IV operator to generate valid codes, followed by the VI operator to produce \textbf{invalid codes}.

\subsubsection{Evolution}
In one evolutionary iteration, we first convert the code into an invalid code via the VI operator, then apply the II operator several times to generate new invalid codes, and finally transform the obtained invalid code back into a valid code using the IV operator.

Since the evolutionary process relies on a fitness function to evaluate codes \cite{Liu2024Evolution}, but invalid codes cannot be directly tested, we define a new fitness function as follows:
\begin{equation}
f(u)=
\begin{cases}
 \frac{\sum \text{eval}_i(u)}{n}, & u \text{ is valid} \\
f(v), & u \text{ is invalid}
\end{cases}
\end{equation}
In this formula, $u$ denotes the valid code or invalid code whose fitness is to be evaluated, $\text{eval}_{i}(u)$ represents the test result of the code on the $i$-th instance, and $v$ denotes the code obtained by repairing $u$ using the LLM. When $u$ is a valid code, its fitness value is defined as the average of the test results over all instances. When $u$ is an invalid code, its fitness is computed by first invoking the LLM to repair it and then evaluating the fitness of the repaired version. Since the repair operation is inherently included in the operations of our proposed operators, it is reasonable to use the repaired code to assess the evolutionary potential of invalid codes.

\subsection{Application in ReEvo}
ReEvo achieves automated heuristic design through reflective evolution, which uses short-term and long-term reflection to guide the LLM in generating new heuristics, yet it is prone to becoming trapped in local optima \cite{Ye2024Reevo}. To alleviate this issue, we enhance the mechanism that guides LLMs to generate novel heuristics based on Abstract Syntax Trees (ASTs). For crossover operations, we first convert codes into ASTs to perform structural crossover, then convert the modified ASTs back into invalid code. The invalid code is then repaired by the LLM under the guidance of short-term reflection. For mutation operations, we convert codes into ASTs to perform structural mutation, restore them to invalid code, and then fix the invalid code using the LLM with the guidance of long-term reflection. In this way, the improved ReEvo can explore a wider search space and better escape local optima.

\subsection{Application in EoH-S}
EoH-S is a paradigm for heuristic set design, which aims to generate highly complementary heuristic sets while adopting the evolutionary operators of EoH \cite{Liu2026Eoh-s}. In this work, we redesign these evolutionary operators based on ASTs. Specifically, heuristic code is first parsed into ASTs, where structural crossover and mutation operations are performed to produce invalid code. The generated invalid code is then fed into prompts to enable the LLM to either repair it directly or generate new valid and diverse codes. By introducing AST-based structural modification, EoH-S gains stronger exploration ability and produces more diverse and complementary heuristic sets, thus improving its overall optimization performance.

\section{EXPERIMENTS}
In this section, we present experimental results of the improved automatic heuristic design algorithm on several representative combinatorial optimization problems, including the Online Bin Packing Problem (OBP) and the Traveling Salesman Problem (TSP). We compare the performance of EoH, ReEvo, and EoH-S before and after the proposed improvements, which validates the superior effectiveness of the AST-based evolutionary operators.

\begin{table*}[htbp]
    \centering
    \caption{Performance comparison on Online Bin Packing (OBP) instances. $n$ and $c$ represent the number of items and the bin capacity, respectively. ``-i'' denotes the enhanced version. Best results are in bold with a light gray background.}
    \label{tbl:obp}
    \begin{tabular}{lccccccc}
        \toprule
        Methods & Training & \multicolumn{6}{c}{Testing ($c=200,500$)} \\
        & (c100) & n1k\_c200 & n5k\_c200 & n10k\_c200 & n1k\_c500 & n5k\_c500 & n10k\_c500 \\
        \midrule
        EoH & 0.0501 & 0.0314 & 0.0223 & 0.0214 & 0.0139 & 0.0071 & 0.0067\\
        EoH-i & \cellcolor{gray!20}\textbf{0.0464} & \cellcolor{gray!20}\textbf{0.0241} & \cellcolor{gray!20}\textbf{0.0187} & \cellcolor{gray!20}\textbf{0.0176} & \cellcolor{gray!20}\textbf{0.0124} & \cellcolor{gray!20}\textbf{0.0063} & \cellcolor{gray!20}\textbf{0.0061}\\
        \midrule
        ReEvo & 0.0442 & 0.0208 & 0.0152 & 0.0150 & 0.0130 & 0.0056 & 0.0053\\
        ReEvo-i & \cellcolor{gray!20}\textbf{0.0348} & \cellcolor{gray!20}\textbf{0.0152} & \cellcolor{gray!20}\textbf{0.0091} & \cellcolor{gray!20}\textbf{0.0086} & \cellcolor{gray!20}\textbf{0.0124} & \cellcolor{gray!20}\textbf{0.0043} & \cellcolor{gray!20}\textbf{0.0043}\\
        \midrule
        EoH-S & 0.0339 & \cellcolor{gray!20}\textbf{0.0182} & 0.0113 & 0.0109 & 0.0124 & 0.0041 & 0.0040\\
        EoH-S-i & \cellcolor{gray!20}\textbf{0.0232} & 0.0340 & \cellcolor{gray!20}\textbf{0.0110} & \cellcolor{gray!20}\textbf{0.0080} & \cellcolor{gray!20}\textbf{0.0124} & \cellcolor{gray!20}\textbf{0.0033} & \cellcolor{gray!20}\textbf{0.0027}\\
        \bottomrule
    \end{tabular}
\end{table*}

\subsection{Experimental Settings and Problems}
\textbf{Experimental Settings:} In our experiments, EoH was set to 10 iterations with a population size of 5, ReEvo was configured with a maximum number of heuristics of 400, and EoH-S was executed for 50 iterations with a population size of 10. \textit{Qwen-Flash} was used as the main large language model, and all the above parameters were kept exactly the same before and after the operator improvement.

\noindent\textbf{Traveling Salesman Problem (TSP):} The Traveling Salesman Problem (TSP) is defined as finding the shortest closed route that visits each city exactly once and returns to the initial starting point. For this problem, we adopt a step-by-step constructive heuristic, which iteratively selects the next city to be visited during path construction. The performance of the proposed algorithm is directly evaluated based on the total path length \cite{Liu2024Evolution}. The training set consists of 128 random instances, where the number of cities ranges from 10 to 200, and the node coordinates are uniformly sampled within the interval $[0,1]$. For the testing phase, three distinct settings (c50, c100, c200, where $c$ represents the number of cities) are employed, with the number of cities ranging from 50 to 200 (consistent with the uniform distribution of the training set). Each method is tested three times independently, and the average results, including token consumption, are recorded for subsequent analysis.

\noindent\textbf{Online Bin Packing (OBP):} The core objective of the Online Bin Packing (OBP) problem is to place sequentially arriving items into bins of fixed capacity with the goal of minimizing the total number of bins used. For this problem, we adopt a designed heuristic that selects a suitable bin for each incoming item, with its performance evaluated by the relative deviation from the lower bound of the optimal bin number \cite{Martello1990Lower}. The training phase uses instance set $\mathcal{I}$ which consists of 128 Weibull-distributed samples with bin capacity 100 and the number of items ranging from 200 to 2000. The testing phase employs six distinct settings (n1k\_c200, n1k\_c500, n5k\_c200, n5k\_c500, n10k\_c200, n10k\_c500, where $n$ and $c$ denote the number of items and bin capacity, respectively) with each method tested three times independently and the average performance metrics reported for analysis \cite{Liu2026Eoh-s}.

\begin{table}[htbp]
    \centering
    \caption{Performance comparison on Traveling Salesman Problem (TSP) instances. $c$ represents the number of cities. ``-i'' denotes the enhanced version. Best results are in bold with a light gray background.}
    \label{tbl:tsp}
    \begin{tabular}{lccccc}
        \toprule
        Methods & Training & \multicolumn{3}{c}{Testing} & Tokens \\
                & (c100) & c50 & c100 & c200 & \\
        \midrule
        EoH & 8.956 & 6.484 & 8.862 & 12.575 & 247010\\
        EoH-i & \cellcolor{gray!20}\textbf{8.900} & \cellcolor{gray!20}\textbf{6.376} & \cellcolor{gray!20}\textbf{8.847} & \cellcolor{gray!20}\textbf{12.517} & \cellcolor{gray!20}\textbf{202529}\\
        \midrule
        ReEvo & 9.180 & 6.484 & 9.224 & 12.988 & 2405312\\
        ReEvo-i & \cellcolor{gray!20}\textbf{8.945} & \cellcolor{gray!20}\textbf{6.474} & \cellcolor{gray!20}\textbf{8.859} & \cellcolor{gray!20}\textbf{12.676} & \cellcolor{gray!20}\textbf{1736673}\\
        \midrule
        EoH-S & 8.809 & 6.405 & \cellcolor{gray!20}\textbf{8.723} & 12.313 & 1493820\\
        EoH-S-i & \cellcolor{gray!20}\textbf{8.762} & \cellcolor{gray!20}\textbf{6.297} & 8.769 & \cellcolor{gray!20}\textbf{12.242} & \cellcolor{gray!20}\textbf{821698}\\
        \bottomrule
    \end{tabular}
\end{table}

\subsection{Experimental Results and Analysis}
\textbf{Experimental Results:} Experimental results demonstrate that the proposed AST-based evolutionary operators consistently improve the performance of all three baseline methods.

Table~\ref{tbl:obp} shows the evaluation of the heuristics designed by the methods on Online Bin Packing (OBP) instances. We can see that on the OBP problem, the improved EoH and ReEvo outperform their pre-improved versions on all instances, while the improved EoH-S is superior to its pre-improved version on all instances except one (n1k\_c200).

Table~\ref{tbl:tsp} shows the evaluation of the heuristics designed by the methods on Traveling Salesman Problem (TSP) instances. As can be seen, for the TSP problem, the optimized EoH and ReEvo algorithms outperform their original iterations on all instances. Meanwhile, the optimized EoH-S algorithm outperforms its original version in all instances except for one (c100). In addition, the token consumption of all optimized algorithms is considerably lower than that of their unoptimized predecessors.

These performance gains validate that AST-based structural operations can effectively expand the search space, help generate stronger heuristics, and reduce token consumption.

\noindent\textbf{Analysis:} The AST-based operator performs crossover and mutation using abstract syntax trees, enabling the algorithm to break free from the knowledge limitations of LLMs, better escape local optima, and discover novel heuristics. Experimental results show that the new operator performs best on ReEvo, likely because ReEvo is an exploitation-oriented algorithm and the new operator provides strong exploration capabilities, leading to optimal synergy; meanwhile, the stochastic nature of node manipulations on ASTs may account for the relatively unstable performance observed across runs. In addition, since the AST operations in the new operator do not require the LLM, the prompts supplied to the LLM are shorter and more focused, resulting in significant savings in token consumption.

In summary, experimental results demonstrate that the proposed AST-based operator design can effectively improve existing automatic heuristic design frameworks. After enhancing EoH, ReEvo, and EoH-S, the improved variants achieve better heuristic algorithms with fewer tokens consumed. Although the improved methods are not sufficiently stable and perform slightly worse on several instances, the overall results are generally in line with our expectations.

\section{CONCLUSION}
This work focuses on Automatic Heuristic Design and proposes an evolutionary operator based on Abstract Syntax Trees (AST). The operator framework mainly includes AST-based destruction and LLM-based repair, and incorporates I-Codes into the evolutionary process. We used this operator to improve existing automatic heuristic algorithms (EoH, ReEvo, EoH-S). Experiments on the Online Bin Packing Problem (OBP) and Traveling Salesman Problem (TSP) show that the improved algorithms are significantly superior to the original ones and consume significantly fewer tokens. This indicates that our operator can effectively expand the search space, find better heuristics, and reduce resource consumption. Future work will focus on directed AST operations and the theoretical improvement of I-Codes. This work not only proposes a new evolutionary operator but also provides a new development idea for LLM-driven automatic heuristic design.


\end{document}